\documentclass[10pt,a4paper]{article}
\usepackage[utf8]{inputenc}
\usepackage[vietnamese]{babel}
\usepackage{times}
\usepackage{amsmath,amsfonts,amssymb}
\usepackage{graphicx}
\usepackage{url}
\usepackage{cite}
\usepackage[table,xcdraw]{xcolor}
\usepackage{microtype}

\usepackage{caption}
\addto\captionsvietnamese{}

\usepackage[paperwidth=20.5cm,paperheight=29.5cm,top=2.2cm,bottom=1.8cm,left=1.8cm,right=1.8cm]{geometry}

\usepackage{fancyhdr}
\fancypagestyle{firstpage}{
    \fancyhf{}
    \fancyfoot[C]{}
}

\usepackage{indentfirst}
\usepackage{setspace}

\usepackage{titlesec}
\titleformat{\section}{\fontsize{10}{12}\bfseries\centering}{\Roman{section}.}{0.5em}{}
\titleformat{\subsection}{\fontsize{10}{12}\bfseries}{\Alph{subsection}.}{0.5em}{}
\titleformat{\subsubsection}{\fontsize{10}{12}\bfseries}{\arabic{subsubsection}.}{0.5em}{}

\begin{document}
\thispagestyle{firstpage}

% Tiêu đề - Arial 14pt, in đậm, chữ hoa
\begin{center}
{\fontfamily{phv}\fontsize{14}{16}\bfseries\selectfont
VietMEAgent: Culturally-Aware Few-Shot Multimodal Explanation for Vietnamese Visual Question Answering
}
\end{center}

\vspace{0.3cm}

% Thông tin tác giả
\begin{center}
{\fontsize{10}{12}\selectfont
Hai-Dang Nguyen\textsuperscript{1}, Minh-Anh Dang\textsuperscript{2}, Minh-Tan Le\textsuperscript{3}, Minh-Tuan Le\textsuperscript{4}\\[0.2cm]
\textsuperscript{1} Faculty Of Information Technology, VNU University of Engineering and Technology, Hanoi, Vietnam\\
\textsuperscript{2} IT-BT Convergence Technology Division, Vietnam-Korea Institute of Science and Technology, Hanoi, Vietnam\\
\textsuperscript{3} TADI Global Lab, TADI Global Company Limited, Hanoi, Vietnam\\
\textsuperscript{4} Faculty of Finance, Banking Academy of Vietnam, Hanoi, Vietnam\\
}
\end{center}

\noindent{\fontsize{9}{11}\bfseries\itshape ABSTRACT}{\fontsize{9}{11}\itshape— 
Contemporary Visual Question Answering (VQA) systems remain constrained when confronted with culturally specific content, largely because cultural knowledge is under-represented in training corpora and the reasoning process is not rendered interpretable to end users. This paper introduces \textit{VietMEAgent}, a multimodal explainable framework engineered for Vietnamese cultural understanding. The method integrates a cultural object detection backbone with a structured program generation layer, yielding a pipeline in which answer prediction and explanation are tightly coupled. A curated knowledge base of Vietnamese cultural entities serves as an explicit source of background information, while a dual-modality explanation module combines attention-based visual evidence with structured, human-readable textual rationales. We further construct a Vietnamese Cultural VQA dataset sourced from public repositories and use it to demonstrate the practicality of programming-based methodologies for cultural AI. The resulting system provides transparent explanations that disclose both the computational rationale and the underlying cultural context, supporting education and cultural preservation with an emphasis on interpretability and cultural sensitivity.}

\vspace{0.2cm}

\noindent{\fontsize{9}{11}\bfseries Keywords}{\fontsize{9}{11}\selectfont— Visual Question Answering, Cultural AI, Explainable AI, Vietnamese Culture, Multimodal Learning.}

\vspace{0.4cm}

\fontsize{10}{12}\selectfont

\section{INTRODUCTION}

Visual Question Answering (VQA) is a central challenge in artificial intelligence that requires a system to jointly process visual and linguistic information to answer questions grounded in images \cite{antol2015vqa,anderson2018bottom}. Despite steady advances, applying VQA to culturally situated scenarios remains difficult. In the Vietnamese context, a model must not only recognize objects, scenes, and relations, but also interpret their cultural meanings, historical connotations, ritual functions, and regional variants. In practice, current systems are often culturally blind: they misrecognize symbols, conflate distinct but visually similar artifacts, or fail to connect a visual cue to its cultural semantics. These issues are exacerbated by a lack of interpretability: black-box predictions obscure the chain of reasoning and hinder verification or correction by non-experts. Language further compounds the problem, as Vietnamese idioms, polysemy, and culturally bound expressions are underrepresented in common benchmarks. Finally, the development of dedicated datasets for Vietnamese Cultural VQA has lagged behind, limiting both modeling and evaluation.

Vietnamese culture spans interlinked domains—from cuisine (e.g., \textit{phở}, \textit{bánh mì}, \textit{bánh xèo}) and architecture (e.g., temples, communal houses, colonial heritage) to attire (e.g., \textit{áo dài} with regional and ceremonial variants), festivals (e.g., Tết, Mid-Autumn, ancestor worship), and folk arts (e.g., traditional instruments, dances, and games). Modeling this landscape requires mechanisms that connect perception with knowledge and that communicate inference in ways users can scrutinize. Recent developments in explainable AI \cite{park2018multimodal,chen2022rex} and few-shot learning with large language models \cite{brown2020language,lu2022learn} suggest new directions: visual program induction can decompose complex questions into verifiable sub-steps \cite{gupta2023visual,suris2023vipergpt}, while open-world recognition expands the space of detectable concepts \cite{minderer2022simple}.

We therefore present \textit{VietMEAgent}, a culturally aware, few-shot multimodal explanation framework. The contributions are fourfold. First, we introduce a Vietnamese Cultural VQA dataset of 28{,}484 annotated samples spanning twelve categories. Second, we propose a multimodal programming approach that couples cultural object detection with structured program generation. Third, we integrate a curated knowledge base of Vietnamese cultural entities to ground both answers and explanations. Fourth, we provide transparent explanations that integrate visual evidence with structured textual reasoning, facilitating educational use and cultural preservation.

\begin{figure}[!h]
\centering
\includegraphics[width=.8\linewidth]{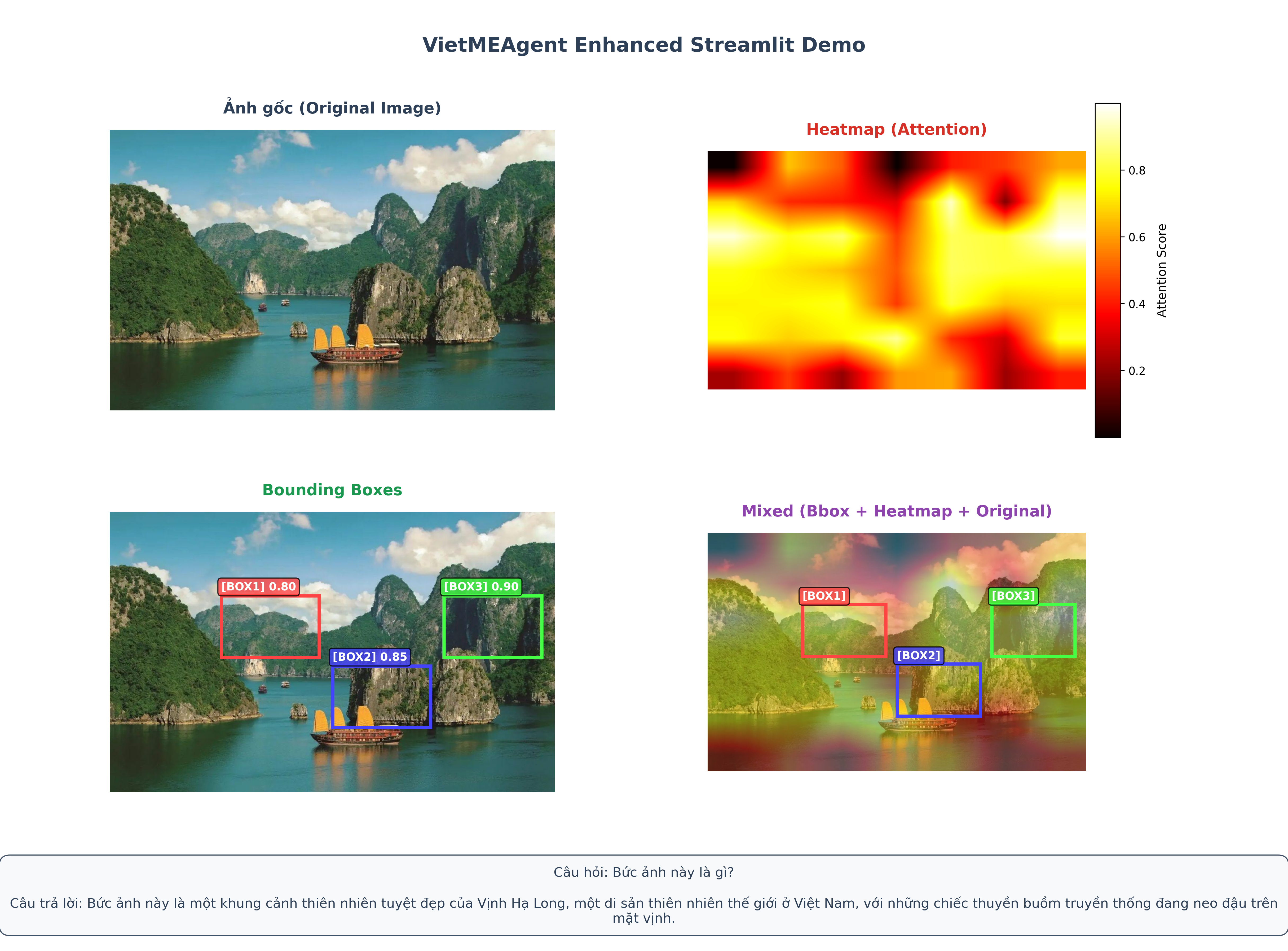}
\caption{VietMEAgent producing multimodal explanations that align visual evidence (e.g., attention heatmaps and bounding boxes) with structured textual rationales. The example illustrates identification of Hạ Long Bay accompanied by contextual cultural information.}
\end{figure}

\section{RELATED WORK}

\subsection{Textual Explanation Methodologies for VQA}

Foundational research emphasized generating textual justifications to accompany answers in order to render the model's reasoning auditable \cite{li2018vqa,park2018multimodal,marasovic2020natural}. Multi-task frameworks that jointly train on VQA and image captioning have been explored \cite{zhou2017neural}, but the decoupling between captions and VQA supervision can yield explanations that are fluent yet not question-specific. Later work attached answer-related captions to VQA instances \cite{li2018vqa} or retrieved auxiliary text to enrich explanations \cite{wen2020multi}. While text-only rationales are accessible and compact, they often elide the visual grounding central to VQA, thereby limiting faithfulness.

\subsection{Multimodal Explanation Frameworks for VQA}

To more closely tie text to perception, multimodal approaches correlate visual attention with explanatory sentences \cite{park2018multimodal,wu2019faithful} or jointly optimize vision–language objectives to improve consistency \cite{chen2022rex,xue2023variational,zellers2019vcr}. For instance, \cite{chen2022rex} maps detected regions into textual slots, though early systems were confined to limited object vocabularies and used rigid templates that reduced readability. Variational formulations have been proposed to strengthen causal alignment between answers and explanations \cite{xue2023variational}. Nevertheless, most methods presuppose large, domain-matched datasets—an assumption that is costly in niche cultural settings. As multimodal LLMs (e.g., GPT-4V \cite{achiam2023gpt}, Gemini-1.5 \cite{reid2024gemini}, Qwen-VL \cite{bai2023qwen}) expand visual reasoning capabilities, lightweight, knowledge-integrated strategies become attractive, especially where cultural precision is paramount.

\subsection{Few-Shot Explanation Methodologies for VQA}

Large language models exhibit strong in-context learning, enabling few-shot rationalization from carefully chosen exemplars \cite{brown2020language,lu2022learn,li2024flexkbqa,lu2024chameleon}. Prior work has shown that prompting with image captions and question–answer exemplars can elicit stepwise reasoning \cite{lu2022learn}. Yet, when cultural content is involved, generic exemplars underperform without explicit domain knowledge. Our approach departs by combining few-shot program induction with a cultural knowledge base and a detector tuned to Vietnamese cultural artifacts. In doing so, we aim to retain the flexibility of few-shot prompting while supplying the precise entities and relations required for culturally faithful explanations.

\section{METHODOLOGY}

\subsection{System Architecture}

VietMEAgent follows a four-stage pipeline that links perception, reasoning, knowledge, and explanation. Given a Vietnamese question and an image, the system first performs cultural object detection to localize salient entities and produce region-level evidence. It then invokes a program generator that translates the natural-language question into an executable, domain-specific program whose functions encode common cultural reasoning patterns. During execution, the program queries a knowledge base to retrieve definitions, historical context, ceremonial significance, and regional variants relevant to the detected entities. Finally, a multimodal explanation module synthesizes visual evidence and structured textual reasoning into a coherent, human-readable account.

\begin{figure}[!h]
\centering
\includegraphics[width=0.96\linewidth]{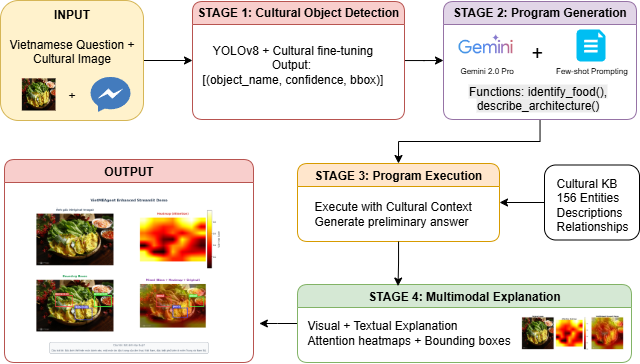}
\caption{Overall architecture. Cultural object detection feeds a program generator that compiles questions into executable steps. Program execution integrates a Vietnamese cultural knowledge base, and an explanation module renders aligned visual and textual rationales.}
\end{figure}

\subsection{Cultural Object Detection Framework}

The detector is built on YOLOv8 and fine-tuned on a culturally curated corpus covering twelve categories, including cuisine, architecture, traditional clothing, and festivals. Training data were collected from Vietnamese cultural repositories and heritage sources, with augmentation strategies that preserve culturally relevant color and texture while improving robustness. The model outputs tuples of the form \texttt{(label, confidence, box)}, for example \texttt{('bánh xèo', 0.92, [x1,y1,x2,y2])}, which provide both semantic labels and spatial evidence for subsequent reasoning and explanation.

\subsection{Program Generation and Execution}

We employ Gemini 2.0 Pro alongside sixteen carefully designed few-shot exemplars to transform Vietnamese cultural questions into structured programs. These programs consist of domain-specific functions such as \texttt{identify\_food}, \texttt{describe\_architecture}, \texttt{explain\_cultural\_significance}, and \texttt{compare\_regional\_variations}. 

The prompts are carefully crafted to reflect Vietnamese linguistic features, including idiomatic expressions and pragmatic particles. They are designed to elicit disambiguation, especially in cases where cultural questions may have multiple valid interpretations. During program execution, visual entities detected in the image are mapped to entries in the cultural knowledge base using both semantic similarity and standardized Vietnamese naming conventions. 

This matching mechanism allows the system to retrieve culturally rich information such as definitions, ceremonial functions, preparation techniques, or historical significance. As a result, both the answer and the explanation are grounded not only in the visual content but also in validated cultural knowledge.

\subsection{Multimodal Explanation Generation}

Explanations are produced by aligning visual and textual channels. On the visual side, attention maps and bounding boxes mark regions most influential to the prediction, making the evidential basis explicit. On the textual side, explanations follow a structured template that first identifies the object or scene, then situates it within Vietnamese cultural context, and finally elaborates on regional variants, historical background, or ceremonial functions as appropriate. A lightweight consistency checker verifies that referenced entities in text correspond to detected regions, and that claims are supported by the knowledge base. The result is a concise yet transparent narrative that users can inspect and contest.

\section{VIETNAMESE CULTURAL VQA DATASET}

\subsection{Construction and Coverage}

We assemble a Vietnamese Cultural VQA dataset comprising 28{,}484 samples across twelve categories representative of core cultural domains. Images are drawn from public sources such as official tourism portals, museum digitizations, cultural heritage repositories, and educational platforms, with careful screening for authenticity and research-appropriate use. Each image is annotated by experts in Vietnamese anthropology, history, or cultural studies. Annotations include culturally correct Vietnamese labels, multiple questions spanning identification, comparison, description, and explanation, and answers that incorporate essential cultural context. A second round of expert review ensures consistency and reduces bias.

The coverage includes everyday and ceremonial cuisine, architectural heritage ranging from religious structures to colonial-era buildings and traditional dwellings, attire encompassing \textit{áo dài} and ethnic costumes, national and local festivals, daily practices and crafts, folk sports and games, traditional transport, handicrafts and decorative arts, musical instruments, and culturally significant landscapes. This breadth is intended to support both recognition and higher-order reasoning within a unified benchmark.

\begin{table}[!h]
\centering
\caption{Dataset Composition by Cultural Category}
\begin{tabular}{|l|c|c|c|}
\hline
\textbf{Cultural Category} & \textbf{Sample Count} & \textbf{Percentage} & \textbf{Complexity Level} \\
\hline
Cuisine  & 2{,}934 & 10.3\% & Low \\
Architecture  & 2{,}991 & 10.5\% & Medium \\
Traditional Clothing  & 2{,}478 & 8.7\% & Low \\
Cultural Festivals  & 2{,}393 & 8.4\% & Medium \\
Daily Life Practices  & 2{,}336 & 8.2\% & High \\
Traditional Sports & 2{,}307 & 8.1\% & High \\
Transportation & 2{,}279 & 8.0\% & High \\
Handicrafts & 2{,}250 & 7.9\% & Very High \\
Miscellaneous Categories & 12{,}516 & 43.9\% & Mixed \\
\hline
\textbf{Total} & \textbf{28{,}484} & \textbf{100.0\%} & \\
\hline
\end{tabular}
\end{table}

\subsection{Quality Assessment and Statistics}

The dataset contains 28{,}484 high-resolution images paired with 91{,}149 questions, yielding an average of approximately 3.2 questions per image. Across the corpus, 156 unique cultural entities are catalogued with expert-validated descriptions to anchor terminology and reduce ambiguity. Question length averages 12.3 words and answers average 15.7 words, reflecting an emphasis on concise yet informative phrasing suited to educational settings. Inter-annotator agreement reaches $\kappa=0.832$, indicating substantial consistency in labeling and reasoning categories, and the overall cultural accuracy approval rate in expert audits is 94.7\%. For research use under appropriate licensing, the dataset is hosted at \url{https://huggingface.co/datasets/Dangindev/viet-cultural-vqa}.

\section{EXPERIMENTAL EVALUATION}

\subsection{Setup and Metrics}

Experiments are conducted on NVIDIA V100 GPUs (32\,GB) using PyTorch~1.12. We evaluate language quality with BLEU-4 \cite{papineni2002bleu}, complement it with METEOR \cite{banerjee2005meteor} and ROUGE-L \cite{lin2004rouge}, and introduce two domain-oriented criteria: Cultural Accuracy, which measures correctness against expert-validated cultural facts, and Explanation Quality, which assesses the coherence and grounding of multimodal rationales. Baselines include BLIP-VQA \cite{li2023blip}, LXMERT \cite{tan2019lxmert}, and a strong large-model variant (PaLM-VQA) representative of instruction-tuned multimodal pipelines.

\subsection{Results and Analysis}

\begin{figure}[!h]
\centering
\includegraphics[width=0.96\linewidth]{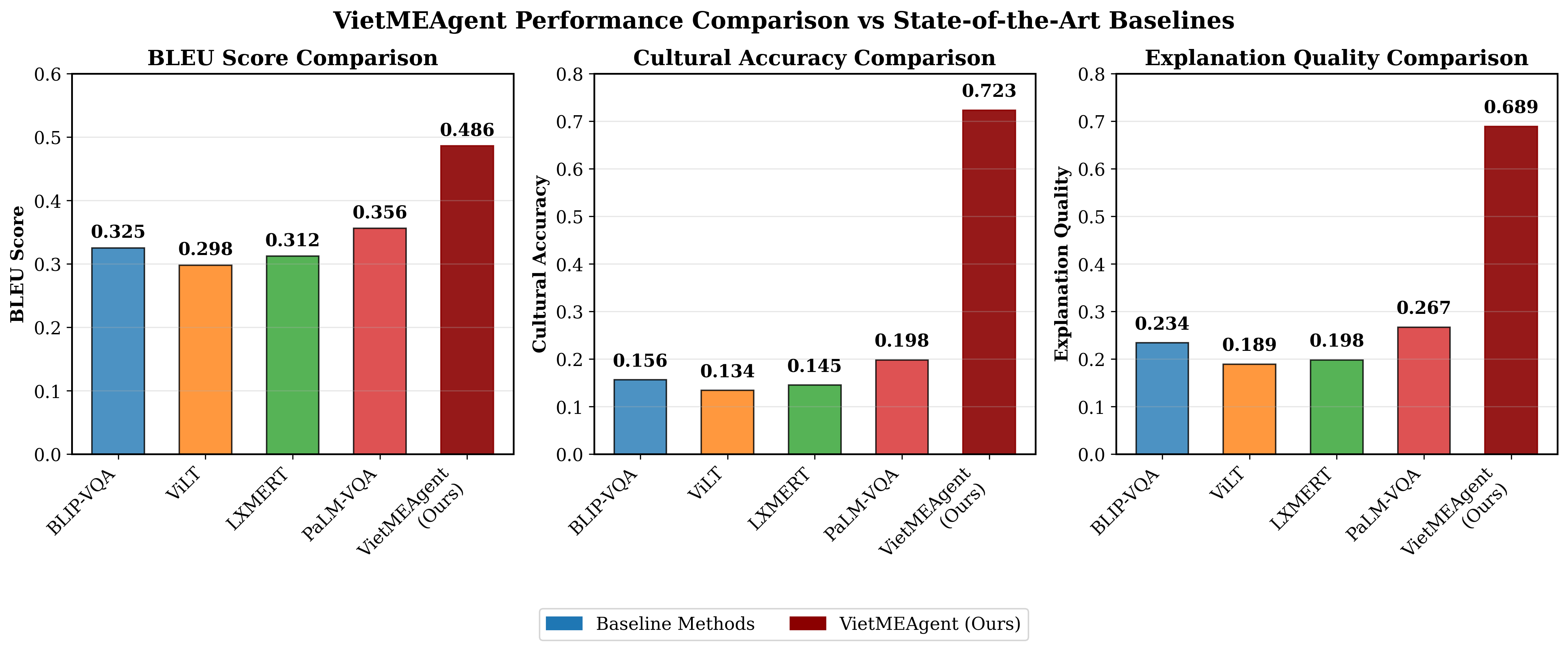}
\caption{Performance comparison across BLEU-4, Cultural Accuracy, and Explanation Quality. VietMEAgent improves both task performance and transparency relative to strong baselines.}
\end{figure}

Quantitatively, VietMEAgent achieves competitive gains over baselines on all metrics, with the largest margins observed in Cultural Accuracy and Explanation Quality, reflecting the benefits of knowledge integration and program-structured reasoning. BLEU-4 scores also improve, suggesting that textual rationales are not only faithful but linguistically fluent. Nevertheless, complex scenes with fine-grained visual distinctions and subtle regional cues remain challenging, indicating room for richer perceptual modules and expanded knowledge entries.

\begin{table}[!h]
\centering
\caption{Quantitative Performance Comparison}
\begin{tabular}{|l|c|c|c|}
\hline
\textbf{Method} & \textbf{BLEU-4} & \textbf{Cultural Accuracy} & \textbf{Explanation Quality} \\
\hline
BLIP-VQA & 0.325 & 0.156 & 0.234 \\
LXMERT & 0.312 & 0.145 & 0.198 \\
PaLM-VQA & 0.356 & 0.198 & 0.267 \\
\hline
\textbf{VietMEAgent} & \textbf{0.486} & \textbf{0.723} & \textbf{0.689} \\
\hline
\end{tabular}
\end{table}

\subsection{Ablation Study}

An ablation analysis isolates the contribution of each module. Removing the cultural knowledge base leads to the most pronounced drop in Cultural Accuracy and Explanation Quality, underscoring the centrality of explicit cultural grounding. Suppressing the visual explanation channel reduces perceived faithfulness even when answers remain correct, while omitting the programming module degrades both accuracy and coherence, likely because the system loses its ability to decompose questions into verifiable sub-steps.

\begin{table}[!h]
\centering
\caption{Ablation Study of VietMEAgent Components}
\begin{tabular}{|l|c|c|c|}
\hline
\textbf{Configuration} & \textbf{BLEU-4} & \textbf{Cultural Accuracy} & \textbf{Explanation Quality} \\
\hline
\textbf{Full Model} & \textbf{0.486} & \textbf{0.723} & \textbf{0.689} \\
w/o Knowledge Base & 0.398 & 0.334 & 0.445 \\
w/o Visual Explanation & 0.456 & 0.678 & 0.534 \\
w/o Programming Module & 0.423 & 0.645 & 0.567 \\
\hline
\end{tabular}
\end{table}

\section{CONCLUSIONS}

We have presented \textit{VietMEAgent}, a few-shot multimodal explanation framework for Vietnamese Cultural VQA that unifies cultural object detection, program-structured reasoning, and knowledge-grounded explanation. The proposed dataset and method jointly advance cultural coverage, interpretability, and task performance. Although complex scenes and subtle regional nuances continue to present difficulties, the results suggest a practical path toward culturally faithful, transparent AI systems. Future work will expand the knowledge base, refine perception for fine-grained recognition, and explore deployment in educational tools and heritage documentation platforms.

\section{ACKNOWLEDGMENTS}

We thank Vietnamese cultural experts for their careful annotations and validation, and we acknowledge the broader research community for constructive feedback during early dissemination. All materials are handled with respect for cultural heritage and ethical guidelines for research.

\section{DATA AVAILABILITY}

The Vietnamese Cultural VQA dataset is available for research purposes at \url{https://huggingface.co/datasets/Dangindev/viet-cultural-vqa} under licenses designed to balance accessibility with responsible use.

\bibliography{references}

@inproceedings{antol2015vqa,
  title={VQA: Visual question answering},
  author={Antol, Stanislaw and Agrawal, Aishwarya and Lu, Jiasen and Mitchell, Margaret and Batra, Dhruv and Zitnick, C Lawrence and Parikh, Devi},
  booktitle={Proceedings of the IEEE international conference on computer vision},
  pages={2425--2433},
  year={2015}
}

@inproceedings{anderson2018bottom,
  title={Bottom-up and top-down attention for image captioning and visual question answering},
  author={Anderson, Peter and He, Xiaodong and Buehler, Chris and Teney, Damien and Johnson, Mark and Gould, Stephen and Zhang, Lei},
  booktitle={Proceedings of the IEEE conference on computer vision and pattern recognition},
  pages={6077--6086},
  year={2018}
}

@inproceedings{chen2022rex,
  title={REX: Reasoning-aware and grounded explanation},
  author={Chen, Shi and Zhao, Qi},
  booktitle={Proceedings of the IEEE/CVF Conference on Computer Vision and Pattern Recognition},
  pages={15586--15595},
  year={2022}
}

@inproceedings{park2018multimodal,
  title={Multimodal explanations: Justifying decisions and pointing to the evidence},
  author={Park, Dong Huk and Hendricks, Lisa Anne and Akata, Zeynep and Rohrbach, Anna and Schiele, Bernt and Darrell, Trevor and Rohrbach, Marcus},
  booktitle={Proceedings of the IEEE conference on computer vision and pattern recognition},
  pages={8779--8788},
  year={2018}
}

@inproceedings{li2018vqa,
  title={VQA-E: Explaining, elaborating, and enhancing your answers for visual questions},
  author={Li, Qing and Tao, Qingyi and Joty, Shafiq and Cai, Jianfei and Luo, Jiebo},
  booktitle={Proceedings of the European Conference on Computer Vision (ECCV)},
  pages={552--567},
  year={2018}
}

@inproceedings{wu2019faithful,
  title={Faithful Multimodal Explanation for Visual Question Answering},
  author={Wu, Jialin and Mooney, Raymond},
  booktitle={Proceedings of the 2019 ACL Workshop BlackboxNLP: Analyzing and Interpreting Neural Networks for NLP},
  pages={103--112},
  year={2019}
}

@inproceedings{xue2023variational,
  title={Variational Causal Inference Network for Explanatory Visual Question Answering},
  author={Xue, Dizhan and Qian, Shengsheng and Xu, Changsheng},
  booktitle={Proceedings of the IEEE/CVF International Conference on Computer Vision},
  pages={2515--2525},
  year={2023}
}

@inproceedings{zellers2019vcr,
  title={From recognition to cognition: Visual commonsense reasoning},
  author={Zellers, Rowan and Bisk, Yonatan and Farhadi, Ali and Choi, Yejin},
  booktitle={Proceedings of the IEEE/CVF conference on computer vision and pattern recognition},
  pages={6720--6731},
  year={2019}
}

@article{brown2020language,
  title={Language models are few-shot learners},
  author={Brown, Tom and Mann, Benjamin and Ryder, Nick and Subbiah, Melanie and Kaplan, Jared D and Dhariwal, Prafulla and Neelakantan, Arvind and Shyam, Pranav and Sastry, Girish and Askell, Amanda and others},
  journal={Advances in neural information processing systems},
  volume={33},
  pages={1877--1901},
  year={2020}
}

@inproceedings{lu2022learn,
  title={Learn to explain: Multimodal reasoning via thought chains for science question answering},
  author={Lu, Pan and Mishra, Swaroop and Xia, Tanglin and Qiu, Liang and Chang, Kai-Wei and Zhu, Song-Chun and Tafjord, Oyvind and Clark, Peter and Kalyan, Ashwin},
  booktitle={Advances in Neural Information Processing Systems},
  volume={35},
  pages={2507--2521},
  year={2022}
}

@article{lu2024chameleon,
  title={Chameleon: Plug-and-play compositional reasoning with large language models},
  author={Lu, Pan and Peng, Baolin and Cheng, Hao and Galley, Michel and Chang, Kai-Wei and Wu, Ying Nian and Zhu, Song-Chun and Gao, Jianfeng},
  journal={Advances in Neural Information Processing Systems},
  volume={36},
  year={2024}
}

@inproceedings{li2024flexkbqa,
  title={FlexKBQA: A flexible LLM-powered framework for few-shot knowledge base question answering},
  author={Li, Zhenyu and Fan, Sunqi and Gu, Yu and Li, Xiuxing and Duan, Zhichao and Dong, Bowen and Liu, Ning and Wang, Jianyong},
  booktitle={Proceedings of the AAAI Conference on Artificial Intelligence},
  volume={38},
  pages={18608--18616},
  year={2024}
}

@inproceedings{gupta2023visual,
  title={Visual programming: Compositional visual reasoning without training},
  author={Gupta, Tanmay and Kembhavi, Aniruddha},
  booktitle={Proceedings of the IEEE/CVF Conference on Computer Vision and Pattern Recognition},
  pages={14953--14962},
  year={2023}
}

@inproceedings{suris2023vipergpt,
  title={ViperGPT: Visual inference via python execution for reasoning},
  author={Sur{\'i}s, D{\'i}dac and Menon, Sachit and Vondrick, Carl},
  booktitle={Proceedings of the IEEE/CVF International Conference on Computer Vision},
  pages={11888--11898},
  year={2023}
}

@inproceedings{minderer2022simple,
  title={Simple open-vocabulary object detection},
  author={Minderer, Matthias and Gritsenko, Alexey and Stone, Austin and Neumann, Maxim and Weissenborn, Dirk and Dosovitskiy, Alexey and Mahendran, Aravindh and Arnab, Anurag and Dehghani, Mostafa and Shen, Zhuoran and others},
  booktitle={European Conference on Computer Vision},
  pages={728--755},
  year={2022},
  organization={Springer}
}

@inproceedings{tan2019lxmert,
  title={LXMERT: Learning Cross-Modality Encoder Representations from Transformers},
  author={Tan, Hao and Bansal, Mohit},
  booktitle={Proceedings of the 2019 Conference on Empirical Methods in Natural Language Processing and the 9th International Joint Conference on Natural Language Processing (EMNLP-IJCNLP)},
  pages={5100--5111},
  year={2019}
}

@inproceedings{li2023blip,
  title={BLIP-2: bootstrapping language-image pre-training with frozen image encoders and large language models},
  author={Li, Junnan and Li, Dongxu and Savarese, Silvio and Hoi, Steven},
  booktitle={Proceedings of the 40th International Conference on Machine Learning},
  pages={19730--19742},
  year={2023}
}

@article{achiam2023gpt,
  title={GPT-4 technical report},
  author={Achiam, Josh and Adler, Steven and Agarwal, Sandhini and Ahmad, Lama and Akkaya, Ilge and Aleman, Florencia Leoni and Almeida, Diogo and Altenschmidt, Janko and Altman, Sam and Anadkat, Shyamal and others},
  journal={arXiv preprint arXiv:2303.08774},
  year={2023}
}

@article{bai2023qwen,
  title={Qwen-VL: A frontier large vision-language model with versatile abilities},
  author={Bai, Jinze and Bai, Shuai and Yang, Shusheng and Wang, Shijie and Tan, Sinan and Wang, Peng and Lin, Junyang and Zhou, Chang and Zhou, Jingren},
  journal={arXiv preprint arXiv:2308.12966},
  year={2023}
}

@article{reid2024gemini,
  title={Gemini 1.5: Unlocking multimodal understanding across millions of tokens of context},
  author={Reid, Machel and Savinov, Nikolay and Teplyashin, Denis and Lepikhin, Dmitry and Lillicrap, Timothy and Alayrac, Jean-baptiste and Soricut, Radu and Lazaridou, Angeliki and Firat, Orhan and Schrittwieser, Julian and others},
  journal={arXiv preprint arXiv:2403.05530},
  year={2024}
}

@article{wen2020multi,
  title={Multi-level knowledge injecting for visual commonsense reasoning},
  author={Wen, Zhang and Peng, Yuxin},
  journal={IEEE Transactions on Circuits and Systems for Video Technology},
  volume={31},
  number={3},
  pages={1042--1054},
  year={2020}
}

@inproceedings{marasovic2020natural,
  title={Natural language rationales with full-stack visual reasoning: From pixels to semantic frames to commonsense graphs},
  author={Marašović, Ana and Bhagavatula, Chandra and Park, Jae Sung and Bras, Ronan Le and Smith, Noah A and Choi, Yejin},
  journal={arXiv preprint arXiv:2010.07526},
  year={2020}
}

@inproceedings{papineni2002bleu,
  title={BLEU: a method for automatic evaluation of machine translation},
  author={Papineni, Kishore and Roukos, Salim and Ward, Todd and Zhu, Wei-Jing},
  booktitle={Proceedings of the 40th annual meeting of the Association for Computational Linguistics},
  pages={311--318},
  year={2002}
}

@inproceedings{banerjee2005meteor,
  title={METEOR: An automatic metric for MT evaluation with improved correlation with human judgments},
  author={Banerjee, Satanjeev and Lavie, Alon},
  booktitle={Proceedings of the ACL workshop on intrinsic and extrinsic evaluation measures for machine translation and/or summarization},
  pages={65--72},
  year={2005}
}

@inproceedings{lin2004rouge,
  title={ROUGE: A package for automatic evaluation of summaries},
  author={Lin, Chin-Yew},
  booktitle={Text summarization branches out},
  pages={74--81},
  year={2004}
}

@inproceedings{zhou2017neural,
  title={More than an answer: Neural pivot network for visual question answering},
  author={Zhou, Yiyi and Ji, Rongrong and Su, Jinsong and Wu, Yongjian and Wu, Yunsheng},
  booktitle={Proceedings of the 25th ACM international conference on Multimedia},
  pages={681--689},
  year={2017}
}
\bibliographystyle{plain}

\end{document}